\documentclass[runningheads]{llncs}
\usepackage{graphicx}
\usepackage{amsfonts}
\usepackage{amsmath}
\begin{document}
\title{Geometric Surface Image Prediction for Image Recognition Enhancement}

\titlerunning{Geometric Surface Image Prediction for Image Recognition Enhancement}
%
%
\author{Tanasai Sucontphunt}
%
%
\institute{National Institute of Development Administration, Thailand }
%
\maketitle              
\begin{abstract}

This work presents a method to predict a geometric surface image from a photograph to assist in image recognition. To recognize objects, several images from different conditions are required for training a model or fine-tuning a pre-trained model. In this work, a geometric surface image is introduced as a better representation than its color image counterpart to overcome lighting conditions. The surface image is predicted from a color image. To do so, the geometric surface image together with its color photographs are firstly trained with Generative Adversarial Networks (GAN) model. The trained generator model is then used to predict the geometric surface image from the input color image. The evaluation on a case study of an amulet recognition shows that the predicted geometric surface images contain less ambiguity than their color images counterpart under different lighting conditions and can be used effectively for assisting in image recognition task.

\keywords{Surface Reconstruction \and Geometric Normal \and Photometric Stereo \and Few-Short Image Recognition \and GAN.}
\end{abstract}
\section{Introduction}

From a photograph, a human can intuitively recognize an object from different lighting. Even objects are very similar to each other, an expert can also tell them apart from a photograph. This is partly because the human can guess the surface of the objects from the photograph. Since there are many self-shadows from ridge and valley in the photograph, a photograph can be ambiguous for a machine to comprehend. Several images with different lighting directions and views are typically required to add variance to the training set. Thus, an image recognition algorithm requires a large photograph dataset to train a model in order to register and recognize the target object. There is still an on-going research on how many and different images are enough for the job~\cite{Shorten2019}. A Few-Shot learning~\cite{Wang2020} attempts to mitigate this issue by re-using the pre-trained model, augmenting image samples, or performing automate learning. For objects with visually different in shapes and colors such as different types of amulets, using pre-trained model with perceptual loss e.g. VGG-16~\cite{simonyan2014} can give very satisfied results. However, if the target objects are very different from the pre-trained dataset, a new dataset of the target objects, which is typically very large, are required to add to the training model. Furthermore, if the target objects are very similar to each other such as human faces or same-shape amulets which can be only differentiated by minor details, the pre-trained model is still needed to be re-trained with a large dataset. For example, to utilize the pre-trained model for very similar faces, the model must be re-trained with several image pairs of the same and different faces using triplet-loss function on Siamese networks~\cite{Koch2015}. Several pair examples are required to induce the feature space so that images of the same object are close together. Nonetheless, for many image recognition tasks, obtaining a large dataset of image pairs at the beginning is very difficult. In this work, a simple Few-Shot image recognition framework is developed where large examples are not require. The recognition is performed by firstly predicting a geometric surface from the input photograph, then extracting features by an existing pre-trained network, and finally recognizing by linear classifiers. To prove the concept, in this work, a pre-trained network model of VGG-16 is used as a perceptual feature extraction. SVM is then used as a linear classifier to recognize the object.

\begin{figure}[htpb]
	\centering
	\includegraphics[scale=0.025]{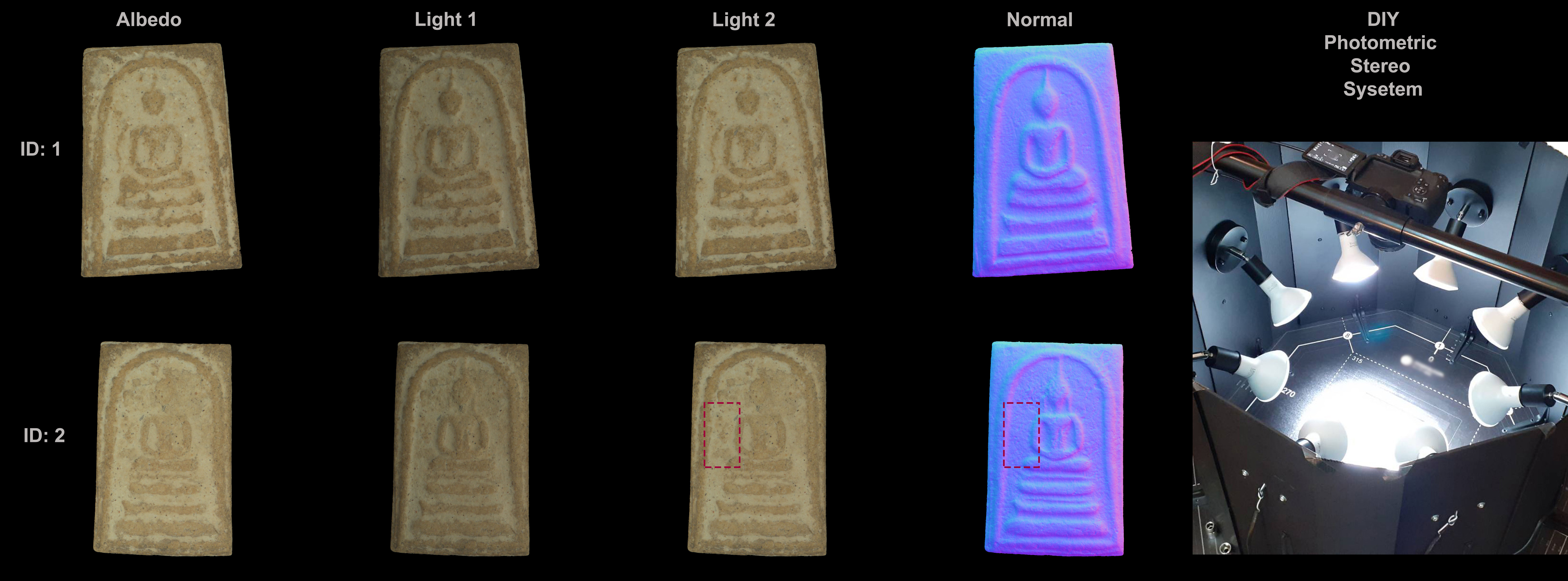}
	\caption{From left to right: Phra Somdej Thai Amulet' albedo image, its color image with lighting direction 1, its color image with lighting direction 2, its normal image. Each row represents different amulet. The red dot boxes show ambiguity surface area from the color image which does not appear in its normal image. Rightmost: a Photometric Stereo Surface Scanning System.}
	\label{fig:phrasomdej_scan}
\end{figure}

An object's surface details i.e. a geometric surface image is used instead of its pure color image in order to reduce the lighting condition dependency. The geometric surface image is a 2D image that contains surface's normal vectors. Thus, the geometric surface image should be a better representation since it does not contain ambiguity from self-shadow i.e. ridge and valley as appearing in the color image. Since each object contain only one geometric image independently from lighting conditions, the geometric surface image can be used as an intrinsic representation of the object. However, capturing object's surface image requires a special equipment e.g. a photometric scanner. For reasonable quality, a DIY photometric stereo can capture finer surface details than using a stereo vision or a depth camera. The photometric stereo uses images from different lighting directions to generate a geometric normal image (a.k.a. geometric surface image) as shown in Fig.~\ref{fig:phrasomdej_scan}. The more sophisticated the system, the higher precision the surface details can be captured. Even though the captured normals may not contain the precise normals but they can be used for image recognition effectively. The geometric normals represent only normal directions from camera view but it does not contain the 3D surface geometry. The normal image can be also used to estimate height map in order to obtain its 3D surface geometry. 

Capturing a geometric normal image by the photometric stereo scanner is not practical for image recognition tasks. In this work, the geometric normal image is synthesized by a generative model from its photograph instead. Firstly, a DIY photometric scanner with eight equally spacing light sources and one high resolution camera is used to create a training dataset (Fig.~\ref{fig:phrasomdej_scan}). The eight images taking from the same camera position but in different light source directions are then used to reconstruct a geometric normal map/image. This reconstruction process takes each image pixel at the same location from eight images with known light source directions to solve for its 3D normal vector at the pixel location. Secondly, to be able to construct a geometric normal image from a single photograph, Generative Adversarial Networks (GAN) is used to map a relation between 2 images from different domains i.e. a geometric normal image and a color photograph image. GAN is a popular method to randomly generate a novel image from example images such as human faces, cartoons, and scenery views.

Lastly, ambiguity evaluation on different lightings of color photographs and predicted normal images are illustrated. Also, Few-Shot image recognition evaluations compare photograph, geometric normal image, and a combination of both for their accuracy and robustness. These evaluations shows that the predicted geometric normal images highly benefit the Few-Shot image recognition in different lighting conditions.

\section{Related Work}

Advanced 3D surface reconstruction systems aim to reassembling information of a target object from cameras or sensors. For example, Smith and Fang~\cite{Smith16} use photometric stereo to reconstruct normal map as well as the height of the surface from photometric ratio. These systems require intensive efforts in sensor setting as well as capturing process. To synthesize a surface image without developing a capturing system, several works combines Bidirectional Reflectance Distribution Function (BRDF) and Photometric Stereo with Convolutional Neural Network (CNN) to generate reflectance maps~\cite{Rematas15}, outdoor surface normal maps~\cite{Holdgeoffroy17}, or Physically Based Rendering (PBR) maps with U-Net~\cite{Deschaintre18}. Multiple images from known illumination directions can also be used to reconstruct their normal maps~\cite{Taniai18} which is similar to this work except known illumination directions is not required in this work.  Many works also focus on using CNN to generate a depth map from a single image such as using global and local layers~\cite{Eigen14} or with Conditional Random Field (CRF)~\cite{Liu14}. Furthermore, coarse normal map~\cite{Trigeorgis17} with albedo~\cite{Sengupta17} or with label~\cite{Eigen15} can also be generated with CNN based techniques from a single image.

Currently, Generative Adversarial Networks (GAN) can be used to translate an image from one domain to another domain with higher details and less artifacts than CNN in general. Also, using GAN in 3D reconstruct gains popularity recently due to its robustness. Special devices can also be used to capture images to train with GAN to reconstruct a surface normal such as using NIR images~\cite{Yoon16} or RGBD images~\cite{Wang16}. Sketches can also be used to reconstruct their surface normals using GAN~\cite{Hudon18}. Face normal map and its albedo map can also be generated with GAN using reconstructed surface as a training set~\cite{Shu17}. In this work, conditional GAN (cGAN)~\cite{Isola17} is used as the main learning model to generate a surface normal from an image. From current GAN family, DualGAN~\cite{Yi17} can also produce similar results to cGAN with less artifacts. However, cGAN is preferred in this work because pixel-to-pixel condition is required to translate from RGB of a color space to its directional XYZ of a normal space precisely. 

\section{Approach}

The main contributions of this work compose of Photometric Data Preparation and GAN-Based Model Training for Normal Map Prediction.

\subsection{Photometric Data Preparation}

Photometric stereo system~\cite{Horn89} is a popular surface reconstruction using images from different illuminations. In this work, photometric stereo system with eight light directions and 45 degrees equally apart is used to capture illumination images as shown in Fig.~\ref{fig:phrasomdej_scan}. This technique reconstructs surface normals from each pixel of all eight illumination images. A surface normal map represents tangent space normals of each captured pixel which is view-dependent to the camera view. In this work, a normal map and its albedo map (pure material colors) are reconstructed without 3D point clouds because the normal image is the main character to be used for image recognition. 

Under Lambertian reflectance assumption (ideal diffuse reflection), surface reflecting light as a color pixel ($I$) on a camera could be derived by $I = k(L \cdot n)$ where $k$ is albedo reflectivity, $L$ is a light direction, and $n$ is surface normal at the point. With known $I$ and $L$ (from light calibration e.g. with mirror ball), at least three light directions (eight in this work) can be used to solve for unknown $kn$ with least-square based method because $I$ is the size of three (RGB). Since $n$ is a unit vector, the magnitude of $kn$ is the $k$ and its unit direction is the $n$. Each $n$ is a 3D vector with ranging value X of [-1, 1], Y of [-1, 1], Z of [-1, 1]. These values will then be used to train in GAN-Based model directly. Normally, a camera is pointing perpendicular to the surface, thus, Z values is limited to [0, 1] which pointing toward the camera. For practical usage, these normal values are stored in UV coordinate of texture map as a normal map. The X,Y,Z values are then stored in the texture map as R of [0, 255], G of [0, 255], B of [128, 255] making the flat surface purple in color e.g. XYZ direction of (0,0,1) is mapping to RGB color of (128,128,255).

\subsection{GAN-Based Model Training for Normal Map Prediction}

\begin{figure}[htpb]
	\centering
	\includegraphics[scale=0.35]{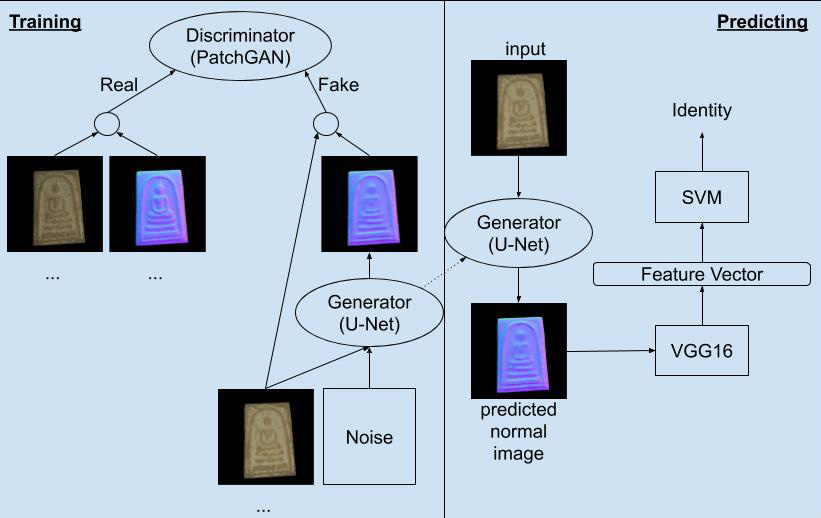}
	\caption{The training and predicting processes in this work.}
	\label{fig:cgan}
\end{figure}

 The goal of using GAN-Based model is to imitate the way photometric stereo calculate for the normal image e.g. with $I = k(L \cdot n)$. Instead of fitting the parameters in least square sense, the GAN-Based model attempts to non-linearly generate each normal image from each color image of different light source. Without specific reconstruction equation, the generated normal image will find relationship between given shading patterns and normal directions implicitly. However, typical GAN-based model generates an image to fit only to its training distribution without exact pixel locations. To give a condition (controllable parameters) for GAN to generate a proper image, conditional Generative Adversarial Networks (cGAN) appends a condition image in the training pipeline to constraint the output image. In this case, the color photographs from different lighting directions are used as condition images to generate their geometric normal image. Figure.~\ref{fig:cgan} (Training) shows an overview of the cGAN training process in this work. The input are photometric-stereo color images and their normal image. The photometric-stereo color image is the condition to feed to both Generator and Discriminator. Since one reconstructed normal image is the result of eight photometric-stereo color images from different light directions, each of eight photometric-stereo color images will be trained with the same normal image. This condition will train the network to capture the reflective nature of the surface. Even though the dataset is small, cGAN can generate a reasonable result without data augmentation since the training set contains highly structure data.

An objective function ($G^*$) for cGAN using in this work is shown in Eq.~\ref{eq:cgan_loss} where $x$ is a color image, $y$ is a normal image, and $z$ is a noise vector. The training loop starts with a Discriminator ($D$), then a Generator ($G$) is trained with the Discriminator's loss together with its loss in iterative. In this work, the PatchGAN, with 32x32 patches, is used as the Discriminator in order to discriminate each patch for fake (0) or real (1). The Discriminator of C64-C128-C256-C512, where C represents Con2D-BatchNorm(except the 1st layer)-LeakyReLU block, is trained directly by detecting if the normal image is fake (predicted normal image from Generator) or real (real normal image) typically with binary-cross-entropy loss. However, in this work, mean-squared-error loss is used instead as it helps the Discriminator learn better in our experiment.

\begin{equation} 
G^* =  \arg\min\limits_{G}\max\limits_{D}(\mathbb{E}_{x,y}[\log D(x,y)] + \mathbb{E}_{x,z}[\log (1-D(x,G(x,z)))]) + \lambda Cos(G)
\label{eq:cgan_loss}	
\end{equation}

To generate fine details of normal maps, U-Net, which is an encoder-decoder with skip connections, is used as the Generator to keep resolutions of the output similar to the input. The U-Net of C64-C128-C256-C512-C512-C512-C512-C512-reverse-back is trained with composite loss of an adversarial loss from the Discriminator and a regularization loss from the generated image. All input images are firstly re-scaled to 512x512 pixels. While training the Generator, the Discriminator will not be trained but it will evaluate for the adversarial loss. To keep generated normal image similar to the real normal image, typically L1 loss is used as a regularization. However, in this work, a cosine similarity ($Cos(G)$) loss (with weight of $\lambda$ = 100) is used instead since the normal image contains only normal vectors. The trainings are optimized using Adam solver with learning rate of 0.0002 and momentum parameters of $\beta_{1}$ = 0.5 and $\beta_{2}$ = 0.999 with batch size of 1 for 200 epochs.

\section{Evaluation}

\begin{figure}[htpb]
	\centering
	\includegraphics[scale=0.07]{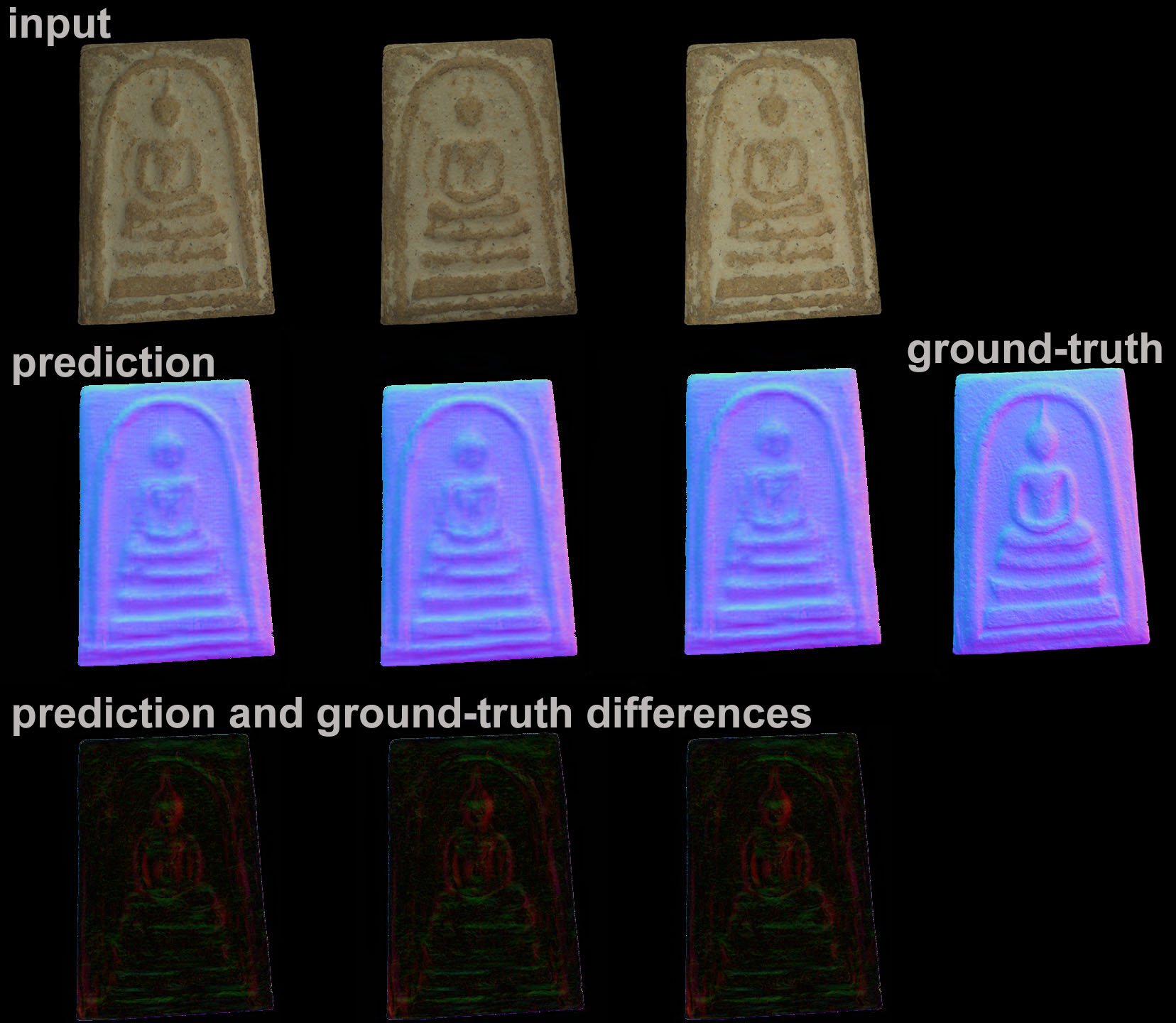}
	\caption{The input color images (from different light directions) and their predicted normal images in comparison. Bottom: the color differences between a ground-truth normal image and its predicted normal images}
	\label{fig:results}
\end{figure}

There are two main evaluations in this work. First, ambiguity evaluation aims to compare ambiguity of color images versus their predicted normal images. Second, Few-Shot image recognition evaluation aims to compare recognition accuracy of color image versus normal image. A particular type of Thai amulets called ``Phra Somdej" is used as the main subject in the evaluations. This type of amulet is typically made from plaster in the shape of tablet as shown in Fig.~\ref{fig:phrasomdej_scan}. There are numerous research works on Thai amulet recognition using CNNs to classify or authenticate~\cite{Mookdarsanit2020} different types of amulets in which feature extraction is sufficient to recognize the amulets effectively~\cite{kompreyarat2015}. However, none of the works focuses on the same type of amulets to identify the exact one. In this work, there are very similar 23 amulets of the same type. To predict all normal images, first 11 amulets are predicted using cGAN trained from another 12 amulets and vice versa. Fig.~\ref{fig:results} shows the input color images comparing to their predicted normal image from cGAN, ground-truth normal image, and the color differences between them. From the figure, the color images are relatively different from lighting directions but their predicted normal images remain resemble to each other. The heat-maps of differences between the predicted normal images and the ground-truth normal image reveal that the predicted normals can mostly represent the real surface structure of the amulet except some crisp silhouettes.

\subsection{Ambiguity Evaluation}

\begin{figure}[htpb]
	\centering
	\includegraphics[scale=0.15]{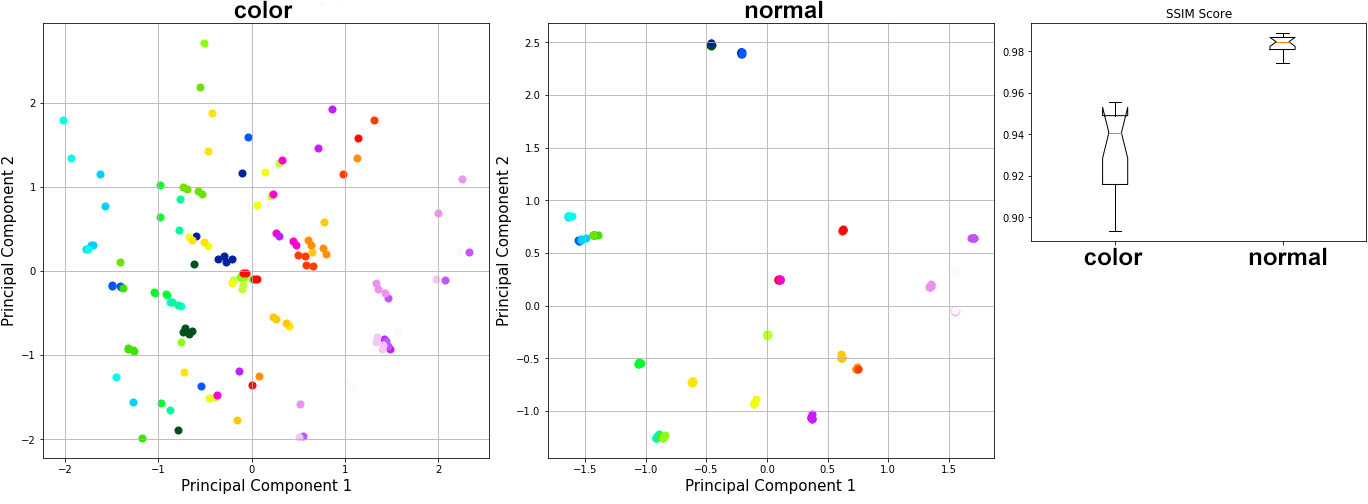}
	\caption{Ambiguity measurement comparing color images and their predicted normal images. Left: color PCA-space. Middle: normal PCA-space. Right: SSIM score box plot.}  
	\label{fig:measurement}
\end{figure}

In this work, the  ambiguity measurement evaluates if the predicted normal image contained less ambiguity than its color image. There are two experiments which are an arrangement visualization and a similarity measurement. To visualize how images are arranged, each type of images are projected to its own domain space using Principal Component Analysis (PCA). The PCA spaces of color and normal images are built from all images of each type. Fig.~\ref{fig:measurement} shows 2D points of principle-component-1 and principle-component-2 of each image where different label colors means different amulets. There are 8 images from different light directions for each amulet. Fig.~\ref{fig:measurement} (left and middle) shows color PCA space and normal PCA space. From this figure, normal images obviously group together closer than color images.

Second, to evaluate if the predicted normal images are similar to each other than their color image inputs, the predicted normal images are measured against their average. As well, the color image inputs of different light directions are measured against their average. The measurement is quantified by Structure Similarity (SSIM) score. Fig.~\ref{fig:measurement} (right) shows a box plot of SSIM score comparing the normal images and the color images. The box plot reveals that the predicted normal images are significantly less sensitive to different light directions. 

\subsection{Few-Shot Image Recognition Evaluation}

From Fig.~\ref{fig:cgan} (Predicting), each amulet has eight images from different light directions. For recognition evaluation, half of them (4 images) is used for registration and another half (4 images) is used for recognition. For registration, the images are transformed to their feature vectors of 4096 dimensions by using VGG-16's second-last fully connected layer. Then, they are trained with SVM to classify for their identities. To add variations, all test images are first transformed by minor rotations and translations before predicted for their normal images. Color images, normal images, and combination of both (by concatenating their feature vectors) are evaluated. To compare which input types are more robust for the recognition, image variations by downgrading of brightness, contrast, color, and blurriness are gradually applied to the test color images before predicting for their normal images. 

\begin{figure}[htpb]
	\centering
	\includegraphics[scale=0.25]{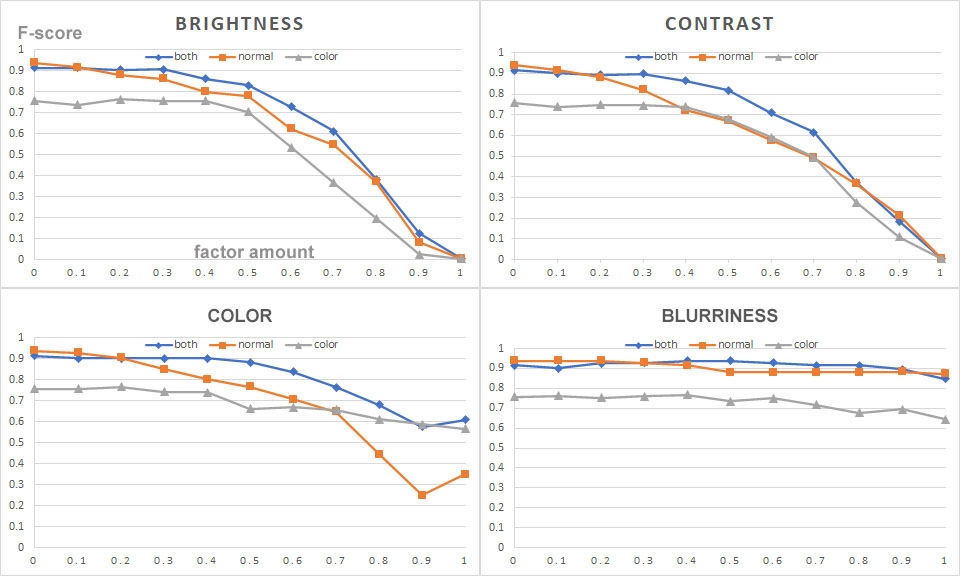}
	\caption{Recognition accuracy by F-score of color, predicted normal, and combination of both images on applying brightness, contrast, color, and blurriness image downgrading.}
	\label{fig:recog}
\end{figure}

Fig.~\ref{fig:recog} shows recognition accuracy by F-score of each image variation with downgrading amounts. The downgrading amount is a real number from 0 to 1.0 where 0 returns original image and 1.0 returns fullest image downgrading, i.e. worsen quality from the original image, which means total black image for brightness, gray image for contrast, black/white image for color, and blurred image for blurriness. From the figure, using color together with its predicted normal for image recognition mostly outperforms color or normal image alone since color and normal features can compensate each other's drawback. However, when the downgrading amounts are less (close to original images), the predicted normal images give higher accuracy than other types. This is because the predicted normal images are accurately predicted from highest quality input color images. For the original images, F-score of color, predicted normal, and combination of both images are 0.75837, 0.93913, and 0.91498 respectively.

\section{Conclusion and Future Work}

In this work, a normal image of target surface is predicted from a color image by using a GAN-Based model which is trained with photometric stereo images captured by a scanner system. The evaluation illustrates that the predicted normal image from this work contain less ambiguity from different illuminations comparing to their color image counterpart and can be used for Few-Shot image recognition effectively. Also, from the evaluation, using both color and predicted normal images together for image recognition give highest accuracy in general. The main reason is that the color information is still important for image recognition. This work does not intend to reconstruct exact surface details. Instead, it tries to create a proper representation for image recognition by predicting its possible surface. 

In the future, more types of target objects such as ancient-heritage objects and more variations of materials will be collected. GAN architecture as well as its loss function can also be adjusted in order to achieve better resolution. Also, this work focuses only on lighting conditions of image ambiguity. However, another important factor which is camera parameters (intrinsic and extrinsic) will need to be addressed. In this case, a 3D model together with its normal map can be used to predict for a better representation. Finally, if there are enough collection of scanned objects, fully automated object identification which robust for real world environments (e.g. taking a photograph from mobile phone under different lighting directions) can be developed.

\section{Acknowledgment}
This project is funded by Research Awards for Complete Research from National Institute of Development Administration.

%
%
%
\bibliographystyle{splncs04}
\bibliography{amulet}
%
\end{document}